\documentclass[10pt,letterpaper]{article}

\input{common-header.sty}
\usepackage[inline]{enumitem}

\author{
François Mazé\textsuperscript{\rm 1},
Faez Ahmed\textsuperscript{\rm 1}
}
\affiliations {
\textsuperscript{\rm 1}Massachusetts Institute of Technology\\
francois.maze@etu.minesparis.psl.eu,
\{fmaze,faez\}@mit.edu
}
\title{Diffusion Models Beat GANs on Topology Optimization}

\begin{document}
\maketitle

\begin{abstract}
Structural topology optimization, which aims to find the optimal physical structure that maximizes mechanical performance, is vital in engineering design applications in aerospace, mechanical, and civil engineering. Generative adversarial networks (GANs) have recently emerged as a popular alternative to traditional iterative topology optimization methods. However, these models are often difficult to train, have limited generalizability, and due to their goal of mimicking optimal structures, neglect manufacturability and performance objectives like mechanical compliance. We propose TopoDiff --- a conditional diffusion-model-based architecture to perform performance-aware and manufacturability-aware topology optimization that overcomes these issues. Our model introduces a surrogate model-based guidance strategy that actively favors structures with low compliance and good manufacturability. Our method significantly outperforms a state-of-art conditional GAN by reducing the average error on physical performance by a factor of eight and by producing eleven times fewer infeasible samples.
By introducing diffusion models to topology optimization, we show that conditional diffusion models have the ability to outperform GANs in engineering design synthesis applications too. Our work also suggests a general framework for engineering optimization problems using diffusion models and external performance with constraint-aware guidance.
We publicly share the data, code, and trained models here:  
https://decode.mit.edu/projects/topodiff/. 

\end{abstract}

\section{Introduction}\label{intro}
Structural topology optimization (TO) of solid structures involves generating the optimal shape of a material by minimizing an objective function, for instance, mechanical compliance, within a given domain and under a given set of constraints (volume fraction, boundary conditions, and loads).
TO is therefore becoming an essential design tool and is now included in most professional design software, such as Autodesk's Fusion 360 and Solidworks.
It is the driving force behind Autodesk's generative design toolset, where designers input design goals into the software, along with parameters such as performance or spatial requirements, materials, manufacturing methods, and cost constraints. The software quickly generates design alternatives.
Most methods to solve TO rely on gradient-based approaches, the most common method being the Solid Isotropic Material with Penalization method~\cite{SIMP1988, SIMP1992}.
Despite their wide adoption, these traditional methods have two major pitfalls: their iterative nature makes them computationally expensive and they may generate non-optimal designs, for example, when penalization and filtering augmentations are used to avoid grayscale pixels in SIMP~\cite{sigmund_review}.

Several deep learning methods have been developed in recent years to improve and speed up the TO process~\cite{Yu2018, Sharpe2019, Nie2021, gantl} by learning from large datasets of optimized structures.
The latest and most promising results were obtained with deep generative models (DGMs) and notably with conditional generative adversarial networks (cGANs) trained for image synthesis, which take as input the boundary conditions and directly generate images of optimized structures. Although popular, most of these models optimize a loss function which does not align with the primary goals of topology optimization --- getting high-performance and feasible structures. They often train for loss functions related to image reconstruction to achieve visual similarity and ignore modeling the physical performance of the generated structures. Most of them produce disconnected, floating material that seriously affects the manufacturability of the generated design. They also suffer from limited generalizability, especially for out-of-distribution boundary conditions. 

We hypothesize that the absence of explicit methods to generate designs with low compliance and good manufacturability causes these issues.
We further hypothesize that the reliance of the optimization objective on the sole cGAN prompts the model to only mimic pixel-wise the ground truth produced by traditional TO methods.
As a result, two images with comparable pixel-wise similarity may still have significantly different performance values.
The absence of explicit external guidance is even more problematic since the ground truth data is not guaranteed to be optimal, as explained above.

This paper introduces TopoDiff, a conditional diffusion-model-based method for TO.
\citet{dhariwal2021diffusion} have shown that diffusion models can outperform GANs for image generation, are easier to train, and are thus more readily adaptable to other tasks. We show that by introducing performance and constraints to diffusion models, they outperform GANs on topology optimization problems too. In addition, the sequential nature of diffusion models makes them compatible with external guidance strategies that assist with performance and feasibility goals.
By creating surrogate models to estimate performance, we thus introduce external guidance strategies to minimize mechanical compliance and improve manufacturability in diffusion models.

Our main contributions include proposing:
(1) TopoDiff --- a diffusion model based end-to-end Topology Optimization framework that achieves an eight-times reduction in average physical performance errors and an eleven-times reduction in infeasibility compared to a state-of-art conditional GAN, (2) a new guidance strategy for diffusion models to perform physical performance optimization with enhanced manufacturability constraint satisfaction, and (3) a generalized framework to solve inverse problems in engineering using diffusion models, when sample feasibility and performance are a high priority.

\section{Background and Related Work}\label{background}

\subsection{Topology Optimization}\label{traditional_TO}
\begin{figure}[h!]
    \centering
    \includegraphics[width = 0.9\columnwidth]{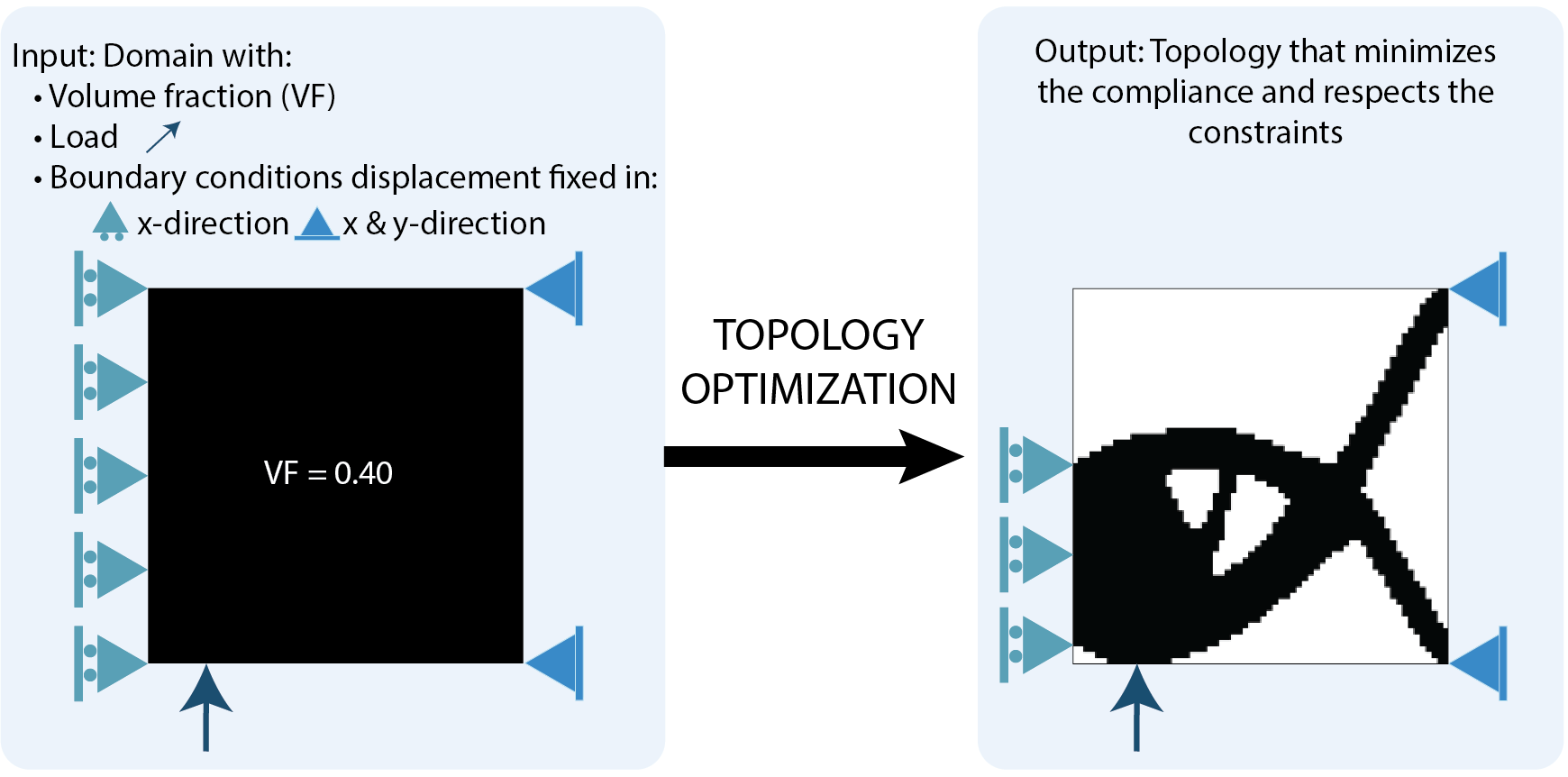}
    \caption{Topology Optimization aims to find the optimal structure that minimizes objectives such as compliance for a given set of load, boundary conditions, and volume fraction.}
    \label{fig:pres_TO}
\end{figure}

Topology Optimization (TO) finds an optimal subset of material $\Omega_{opt}$ included in the full design domain $\Omega$ under a set of displacement boundary conditions and loads applied on the nodes of the domain and a volume fraction condition.
A structure is optimal when it minimizes an objective function, such as mechanical compliance, subject to constraints.
Fig. \ref{fig:pres_TO} summarizes the principle of TO.

Traditional TO methods rely on Finite Elements Analysis (FEA) using gradient-based~\cite{SIMP1988} or gradient-free methods~\cite{ahmed2013constructive}.
One popular gradient-based method is the Solid Isotropic Material with Penalization (SIMP) method~\cite{SIMP1992}.

SIMP associates every element of the mesh with a \textit{continuous} density to perform gradient-based methods~\cite{99code_matlab}.
However, because intermediary densities make no physical sense, SIMP uses a penalization factor to encourage binary densities.

This penalization strategy (with $p > 1$) is efficient but introduces non-convexity in the objective function, as stated by \citet{sigmund_review}.
As a result, SIMP is likely to converge towards local optima.
Other techniques to encourage binary densities include filters, but they also introduce non-convexity~\cite{sigmund_review}.

\subsection{Deep Learning for Topology Optimization}\label{DL_for_TO}
Traditional TO methods are often slow due to the iterative FEA steps they include~\cite{computational_time_trad_TO}.
Many deep learning methods~\cite{10.1115/1.4053859, Guo2018, Lin2018, Sosnovik2019} have recently been developed to improve the speed and quality of topology generation or address issues such as non-convexity.

A group of deep learning approaches, to which our work belongs, consists in proposing an end-to-end TO framework from the constraints to the optimal topology~\cite{oh2019, Sharpe2019, 10.1007/s00158-020-02748-4, parrott2022multi}. Below, we review a few representative works. \citet{Yu2018} propose an iteration-free method, which predicts a low-resolution solution with a CNN encoder-decoder, which is then passed into a GAN to increase the resolution.
In line with the work by \citet{Rawat2019}, \citet{Li2019} use two GANs to solve the TO problem and then predict the refined structure at high resolution. 
\citet{Sharpe2019} introduce conditional GANs as a means of generating a compact latent representation of structures resulting from TO.
Their work is improved by \citet{Nie2021}, who greatly extend the generalizability in their model named TopologyGAN, which is trained on more diverse conditions and uses physical fields as input to represent loads and boundary conditions.
In parallel, \citet{Zhang2022} develop a U-Net to perform TO for improved generalization.
These promising models nevertheless show a limited generalization ability, notably regarding boundary conditions outside the training distribution.
They are also both prone to the problem of disconnected material.
To solve this issue, \citet{gantl} propose a conditional GAN architecture that includes a topological measure of connectivity in its loss function.
Their results seem to improve generalizability and connectivity; however, they set the volume fraction to a constant value, limiting the problem's scope.

It is crucial to note that none of these methods explicitly include a process to minimize compliance, which is the goal of TO. The minimization of compliance is expected indirectly through the GAN training, which is challenging to control. 
As incorporating some measure of predicted structural performance inside a conditional model seems necessary, we propose explicit guidance methods in diffusion models for low-compliance and good-feasibility structures.

\subsection{Diffusion Models}\label{diffusion_models}
Diffusion models are a new type of deep generative models (DGMs) introduced by \citet{dickstein2015}.
They have received much attention recently because \citet{dhariwal2021diffusion} showed that diffusion models outperform GANs for image synthesis.
Diffusion models are increasingly being applied to various fields: image generation~\cite{image_gen}, segmentation~\cite{image_segmentation}, image editing~\cite{image_editing}, text-to-image~\cite{text_to_image1, text_to_image2}, etc.

The idea behind diffusion models is to train a neural network to reverse a noising process that maps the data distribution to a white noise distribution.
The forward noising process, which is fixed, consists of progressively adding noise to the samples following the Markov chain:
\begin{equation}
q(x_t|x_{t-1}) = \mathcal{N}(x_t; \sqrt{\alpha_t}x_{t-1}, (1-\alpha_t)I)    
\end{equation}
where $(\alpha_t)_{t=1}^{T}$ is a variance schedule.

To reverse this noising process, we approximate the true posterior with the parametric Gaussian process:
\begin{equation}
p_{\theta}(x_{t-1}|x_t) = \mathcal{N}(\mu_{\theta}(x_t), \Sigma_{\theta}(x_t)).
\label{gaussian_process}
\end{equation}

We then generate new data by sampling an image from $\mathcal{N}(0, I)$ and gradually denoising it using Eq. \ref{gaussian_process}.

Training a diffusion model, therefore, involves training two neural networks, $\mu_{\theta}(x_t)$ and $\Sigma_{\theta}(x_t)$, to predict the mean and the variance of the denoising process respectively.
Let us note that \citet{ho2020denoising} showed that $\Sigma_{\theta}(x_t)$ might be fixed to a constant instead of being learned.

\subsection{Guidance Methods in Diffusion Models}\label{guidance_methods}
In many machine learning applications, a model is expected to generate samples conditioned on some input conditions. For example, popular text-to-image models such as DALL-E are conditioned on text input. Researchers have developed a few guidance methods to perform conditional image generation, such as including class labels when the model tries to generate an image corresponding to a given class.
\subsubsection{Including conditioning information inside the denoising networks}\label{cond_info_in_network}
A method to condition a diffusion model consists in adding the conditioning information (for example, a class label) as an extra input to the denoising networks $\mu_{\theta}$ and $\Sigma_{\theta}$.
In practice, the conditioning information can be added as an extra channel to the input image.
Similarly, \citet{dhariwal2021diffusion} suggest adding conditioning information into an adaptive group normalization layer in every residual block.
\subsubsection{Classifier guidance}\label{classifier_guidance}
Additional methods have been developed to guide the denoising process using classifier output.
In line with \citet{dickstein2015} and \citet{sdesong}, who have used external classifiers to guide the denoising process, \citet{dhariwal2021diffusion} introduce \textit{classifier guidance} to perform class-conditional image generation.
In \textit{classifier guidance}, a separate classifier is trained on noisy data (with different levels of noise) to predict the probability $p_{\phi}(y|x_t)$ that an image $x_t$ at noise level $t$ corresponds to the class $y$.
Let $p_{\theta}(x_t|x_{t+1})$ be an unconditional reverse noising process. 
Classifier guidance consists of sampling from:
\begin{equation}
    p_{\theta, \phi}(x_t|x_{t+1}, y) = Z p_{\theta}(x_t|x_{t+1}) p_{\phi}(y|x_t)
\end{equation}
instead of $p_{\theta}(x_t|x_{t+1})$, where $Z$ denotes a normalizing constant.
Under reasonable assumptions, \citet{dhariwal2021diffusion} show that sampling from $p_{\theta, \phi}(x_t|x_{t+1}, y)$ is equivalent to perturbing the mean with the gradient of the probability predicted by the classifier. Specifically, the perturbed mean is:
\begin{equation}
    \hat{\mu_{\theta}}(x_t) = \mu_{\theta}(x_t) + s \Sigma_{\theta}(x_t) \nabla_{x_t} \log p_{\phi}(y|x_t)
\end{equation}
where $s$ is a scale hyperparameter that needs to be tuned. A variant called classifier-free guidance is proposed by \citet{ho2021classifierfree}.
This technique is theoretically close to classifier guidance but does not require training a separate classifier on noisy data.

However, none of these methods provide guidance for both continuous values (such as performance obtained from regression models) and discrete values (such as class labels obtained from classification models), which is important for TO to achieve feasible, high-performing samples. To overcome these issues, we propose a regressor and classifier guidance strategy that penalizes low-compliance and infeasible structures at every step.

\section{Method}\label{method}

\subsection{Architecture and General Pipeline}\label{architecture}
TopoDiff's diffusion architecture consists of a UNet~\cite{UNet}-based denoiser at every step with attention layers. We add conditioning to this architecture by including information on constraints and boundary conditions as additional channels to the input image given to the denoiser, as shown in Figure 2.
The UNet model uses these extra channels as additional information to denoise the first channel of the input in a way that respects the constraints and is optimal for the given boundary conditions.
Similarly to TopologyGAN~\cite{Nie2021}, we use physical fields, namely strain energy density and von Mises stress, to represent constraints and boundary conditions.
The physical fields are computed using a finite element method \cite{solidspy} and help avoid the sparsity problem caused by raw constraints and boundary condition matrices.
The final input to our conditional diffusion model has four channels representing the volume fraction, the strain energy density, the von Mises stress, and the loads applied to the domain's boundary. 

\begin{figure*}
  \centering
  \includegraphics[scale = 0.35]{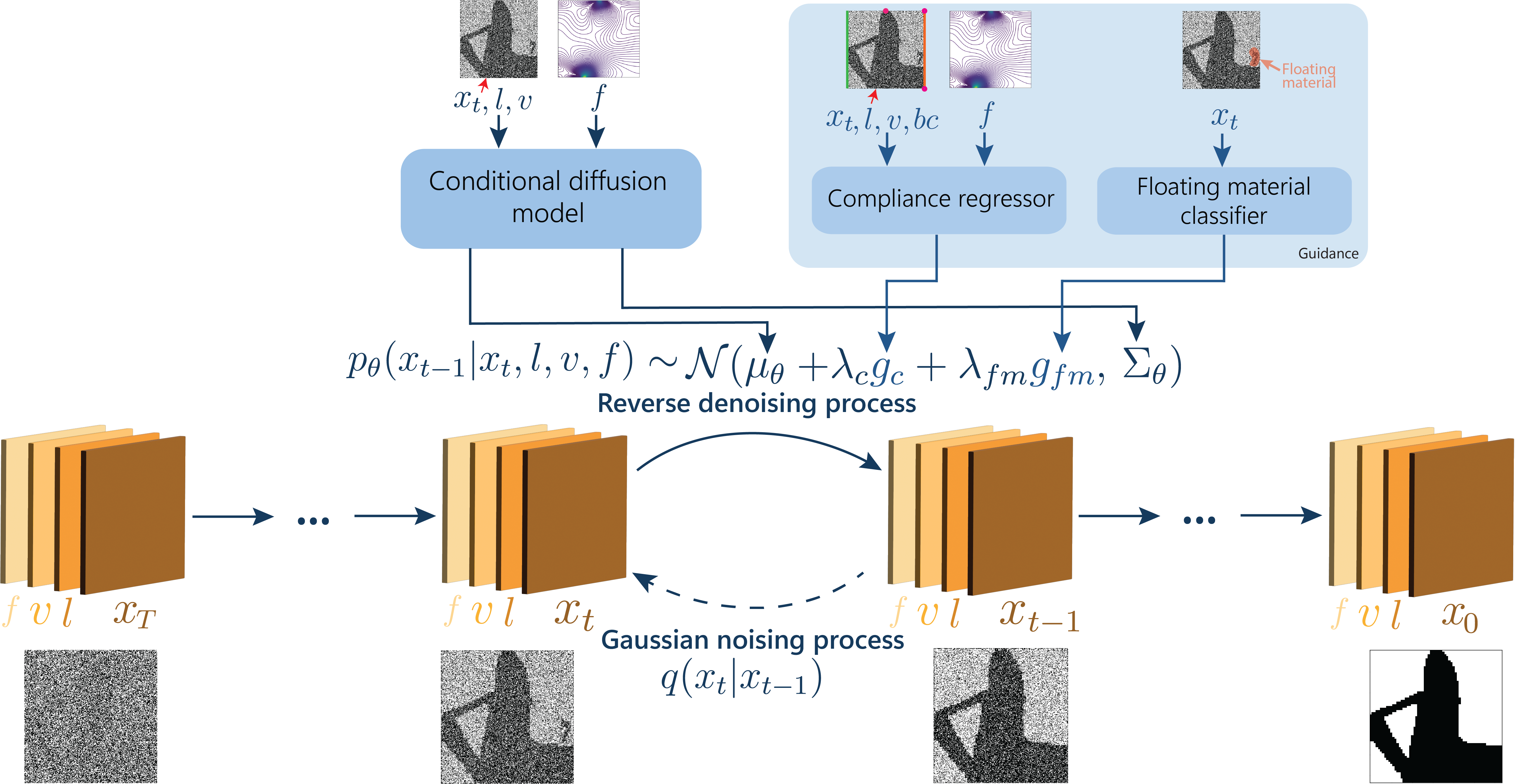}
  \caption{TopoDiff: Proposed constrained guided conditional diffusion model architecture for TO. $(x_t)_{t=0,\cdots,T}$ is the gradually denoised topology; $g_c$ and $g_{fm}$ are the guidance gradients; $l$ is the load applied, represented with an arrow on the topology; $v$ is the volume fraction; $f$ are the physical fields, and $bc$ are the boundary conditions, represented with color lines and dots.}
  \label{fig:full_diff_model}
\end{figure*}
\subsection{Minimizing compliance} \label{compliance_guidance}
Most deep learning models used for TO rely on minimizing the pixel-wise error between the output topology and the ground truth obtained with traditional methods.
For instance, the reference model TopologyGAN~\cite{Nie2021} attempts to mimic the ground truth topology and is encouraged to do so by the L2 loss function of its generator.
GANs for TO are often evaluated using mean absolute error (MAE) between the ground truth topology and the topology predicted by their model.
However, we hypothesize that setting the minimization of a pixel-wise error as an objective does not properly address the aim of TO: generate manufacturable structures that minimize mechanical compliance. We pose this hypothesis for two main reasons:
\begin{enumerate}
    \item The topology used as ground truth may be sub-optimal due to penalization factor and filters (Sec. \ref{traditional_TO});
    \item A small pixel-wise error is compatible with a large compliance error if the material is missing at critical places.
\end{enumerate}
Without additional guidance, our conditional diffusion model is prone to the same problem.
To solve that issue, we introduce a new type of guidance called \textit{regressor guidance}, inspired by the \textit{classifier guidance} used by \citet{dhariwal2021diffusion}.

Consider a conditional diffusion model as presented in Sec. \ref{architecture}:  $p_{\theta}(x_t|x_{t+1}, v, f, l)$, where $v$ is the \textit{volume fraction}, $f$ are the physical fields (strain energy density and von Mises stress) and $l$ are the loads applied. \textit{Regressor guidance} consists in sampling each transition according to:
\begin{multline}
    p_{\theta, \phi}(x_t|x_{t+1}, v, f, l, bc) =\\Z p_{\theta}(x_t|x_{t+1}, v, f, l) e^{-c_{\phi}(x_{t}, v, f, l, bc)}
\end{multline}
where $c_{\phi}$ is a surrogate neural network predicting the compliance of the topology under given constraints, $bc$ are the boundary conditions applied, and $Z$ is a normalizing constant.
It is worth noting that $c_{\phi}$ must be able to predict compliance on noisy images of structures.
To perform this task, we use the encoder of a UNet architecture modified for regression values.

We show in Sec. \ref{maths} that under simple assumptions, adding \textit{regressor guidance} simply amounts to shifting the mean predicted by the diffusion model by $- \Sigma \nabla_{x_t} c_{\phi}(x_{t}, v, f, l, bc)$ where $\Sigma$ is the variance of the Gaussian distribution representing $p_{\theta}(x_t|x_{t+1}, v, f, l)$. This method thus modifies the distribution according to that which we sample at each step by penalizing structures with high compliance. The resulting algorithm is Alg. \ref{alg:reg_guidance}.

\begin{algorithm}[h]
\caption{Regressor guidance for TO, given a conditional diffusion model $(\mu_{\theta}(x_t|x_{t+1}, v, f, l),\Sigma_{\theta}(x_t|x_{t+1}, v, f, l))$ and a regressor $c_{\phi}(x_{t}, v, f, l, bc)$.}\label{alg:reg_guidance}
\begin{algorithmic}
\Require $v, l, bc$ \Comment{Volume, loads and boundary conditions}
\Require $f$ \Comment{Physical fields}
\Require $\lambda_{c}$ \Comment{Regressor gradient scale}
\State $x_T \gets \textrm{sample from } \mathcal{N}(0,\,I)$
\For{$t$ from $T$ to 1}
    \State $\mu, \Sigma \gets \mu_{\theta}(x_t|x_{t+1}, v, f, l),\Sigma_{\theta}(x_t|x_{t+1}, v, f, l)$
    \State $x_{t-1} \gets \textrm{sample from } \mathcal{N}(\mu - \lambda_{c} \Sigma \nabla_{x_t} c_{\phi}(x_{t}, v, f, l, bc)|_{x_t = \mu},\,\Sigma)$
\EndFor
\Return $x_0$
\end{algorithmic}

\end{algorithm}

\subsection{Avoiding Floating Material}\label{fm_guidance}
Disconnected pixels in predicted structures are a serious problem because this phenomenon leads to floating material and therefore affects the manufacturability of the predicted topology.
This problem is generally ignored in deep learning models for TO, and is notably not considered by the pixel-wise error because a small pixel-wise error is compatible with the presence of floating material.

Similar to what has been exposed in Sec. \ref{compliance_guidance}, we further modify the sampling distribution at each step by penalizing structures that contain floating material. To do so, we train a classifier $p_{\gamma}$ that returns the probability that the topology does \emph{not} contains floating material.
We then use this classifier to perform \textit{classifier guidance}, as introduced in Sec. \ref{guidance_methods}. Eventually, this amounts to shifting the mean predicted by the diffusion model by $+ \Sigma \nabla_{x_t} \log p_{\gamma}(x_{t})$.

\subsection{Combining Guidance Strategies}\label{combining_guidance}
Our model ultimately consists of one conditional diffusion model $p_{\theta}(x_t|x_{t+1}, v, f, l)$ and two surrogate models used for guidance when sampling: $c_{\phi}(x_{t}, v, f, l, bc)$ for compliance and $p_{\gamma}(x_{t})$ for floating material.

One challenge is to find a way to combine these two guidance strategies. To combine them, we sample at every step according to:
\begin{multline}
p_{\theta, \phi, \gamma}(x_t|x_{t+1}, v, f, l, bc) =\\Z p_{\theta}(x_t|x_{t+1}, v, f, l) e^{-c_{\phi}(x_{t}, v, f, l, bc)} p_{\gamma}(x_{t}).
\end{multline}

This amounts to shifting the mean predicted by the diffusion model by:
\begin{multline}
- \lambda_{c} \Sigma \nabla_{x_t} c_{\phi}(x_{t}, v, f, l, bc) + \lambda_{fm} \Sigma \nabla_{x_t} \log p_{\gamma}(x_{t})
\end{multline}
where $\lambda_{c}$ and $\lambda_{fm}$ are gradient scale hyperparameters.

However, as is, this approach has two pitfalls:
\begin{enumerate*}
    \item The gradients are always computed at the same point $\mu$ (the mean predicted by the diffusion model), even though this mean is shifted by the previous guidance strategy; and
    \item The gradients are computed at every denoising step, even if we might want to favor one guidance over the other at a given denoising step.
\end{enumerate*}
We modify the point at which the second gradient is computed to address these issues by considering the shift induced by the previous gradient.
In addition, we determine a maximum level of noise (MLN) beyond which the \textit{classifier/regressor guidance} should not be included for every classifier and regressor. 
We then introduce \textit{classifier/regressor guidance} only if the image is denoised enough to have a noise level below the MLN of the given classifier or regressor.

The final guidance algorithm resulting from the combination of these guidance strategies is Alg. \ref{alg:guidance_strategy}.
Fig. \ref{fig:full_diff_model} also summarizes the overall architecture.

\begin{algorithm}
\caption{Guidance strategy for TO using Conditional Diffusion Model.}\label{alg:guidance_strategy}
\begin{algorithmic}
\Require $v, l, bc$ \Comment{Volume, loads and boundary conditions}
\Require $f$ \Comment{Physical fields}
\Require $\lambda_{c}, \lambda_{fm}$ \Comment{Regressor and classifier gradient scale}
\Require $MLN_{c}, MLN_{fm}$ \Comment{Maximum levels of noise}
\State $x_T \gets \textrm{sample from } \mathcal{N}(0,\,I)$
\For{$t$ from $T$ to 1}
    \State $\mu, \Sigma \gets \mu_{\theta}(x_t|x_{t+1}, v, f, l),\Sigma_{\theta}(x_t|x_{t+1}, v, f, l)$
    \If{$t < MLN_{fm}$}
        \State $\mu \gets \mu + \lambda_{fm} \Sigma \nabla_{x_t} \log p_{\gamma}(x_{t})|_{x_t=\mu}$
    \EndIf
    \If{$t < MLN_{c}$}
        \State $\mu \gets \mu - \lambda_{c} \Sigma \nabla_{x_t} c_{\phi}(x_{t}, v, f, l, bc)|_{x_t=\mu}$
    \EndIf
    \State $x_{t-1} \gets \textrm{sample from } \mathcal{N}(\mu,\,\Sigma)$
\EndFor
\Return $x_0$
\end{algorithmic}
\end{algorithm}

It should be noted that we also considered applying \textit{regressor} and \textit{classifier guidance} for other constraints (volume, load position), but the conditional diffusion model already sufficiently respects these constraints and makes guidance unnecessary.

\subsection{Mathematical Motivations for \textit{Regressor Guidance}}\label{maths}
Similarly to \citet{dhariwal2021diffusion} about \textit{classifier guidance}, we show in this section the mathematical motivations behind \textit{regressor guidance}, and in particular, we prove that adding \textit{regressor guidance}, \textit{i.e.,} sampling each transition according to the modified distribution, amounts to a shift in the mean predicted by the diffusion model.

Let $p_{\theta}(x_t|x_{t+1}, v, f, l)$ be our conditional diffusion model. \textit{Regressor guidance} consists in sampling according to:
\begin{multline}\label{weighted_distribution}
    p_{\theta, \phi}(x_t|x_{t+1}, v, f, l, bc) =\\Z p_{\theta}(x_t|x_{t+1}, v, f, l) e^{-c_{\phi}(x_{t}, v, f, l, bc)}
\end{multline}
where $Z$ is a normalizing constant and all variables are the ones defined in Sec. \ref{compliance_guidance}.

Let $\mu$ and $\Sigma$ be the mean and variance of the Gaussian distribution representing $p_{\theta}(x_t|x_{t+1}, v, f, l)$.
\begin{multline}
    \log p_{\theta}(x_t|x_{t+1}, v, f, l) =\\-\frac{1}{2}(x_t-\mu)^{T}\Sigma^{-1}(x_t-\mu)+C
\end{multline}
where $C$ is a constant.

By doing a Taylor expansion of the regressor predicting the compliance $c_{\phi}(x_t, v, f, l, bc)$, we obtain:
\begin{multline}\label{taylor}
    c_{\phi}(x_t, v, f, l, bc) \approx c_{\phi}(\mu, v, f, l, bc)\\+ (x_t-\mu) \nabla_{x_t}c_{\phi}(x_t, v, f, l, bc)|_{x_t=\mu}.
\end{multline}
In Eq. \ref{taylor}, we neglect the terms of second order and above because we make the assumption that the curvature of $c_{\phi}(x_t)$ is low compared to $\Sigma^{-1}$, to which it will be summed in Eq. \ref{sum_taylor_gauss}.
This assumption is reasonable in the limit of infinite diffusion steps, where $||\Sigma|| \rightarrow 0$, as stated by \citet{dhariwal2021diffusion}.

Hence, Eq. \ref{taylor} can be rewritten $c_{\phi}(x_t, v, f, l, bc) \approx (x_t-\mu) g_c + D$, where $g_c$ is the gradient of the compliance regressor evaluated in $\mu$ and $D$ is a constant.

This gives:
\begin{multline}\label{sum_taylor_gauss}
    \log (p_{\theta}(x_t|x_{t+1}, v, f, l) e^{-c_{\phi}(x_{t}, v, f, l, bc)}) \approx \\ -\frac{1}{2}(x_t-\mu)^{T}\Sigma^{-1}(x_t-\mu)-(x_t-\mu)g_c + C + D.
\end{multline}

Hence:
\begin{multline}\label{final_eq}
    \log (p_{\theta}(x_t|x_{t+1}, v, f, l) e^{-c_{\phi}(x_{t}, v, f, l, bc)}) \approx \\ -\frac{1}{2}(x_t-\mu+\Sigma g_c)^{T}\Sigma^{-1}(x_t-\mu+\Sigma g_c)+\frac{1}{2}g_{c}^{T} \Sigma g_c + C +D.
\end{multline}

The last three terms of Eq. \ref{final_eq} are all constant and are encapsulated in the normalizing constant $Z$ from Eq. \ref{weighted_distribution}. Therefore, we have shown that $p_{\theta, \phi}(x_t|x_{t+1}, v, f, l, bc)$ can be approximated by a Gaussian with a mean shifted by $-\Sigma g_c$.

\section{Empirical Evaluation}\label{empirical_evaluation}
We created three datasets to train the proposed models, which we make public to provide a standard benchmark for future research in this area.

\subsection{Dataset}\label{dataset}
The main dataset used to train, validate and test TopoDiff consists of 33000 64x64 2D images corresponding to optimal structures for diverse input conditions.
Every data sample contains six channels:
\begin{enumerate}
    \item The first channel is the black and white image representing the optimal topology;
    \item The second channel is uniform and includes the prescribed volume fraction;
    \item The third channel is the von Mises stress of the full domain under the given load constraints and boundary conditions, defined as $\sigma_{vm} = \sqrt{\sigma_{11}^{2} - \sigma_{11}\sigma_{22} + \sigma_{22}^{2} + 3\sigma_{12}^{2}}$;
    \item The fourth channel is the strain energy density of the full domain under the given load constraints and boundary conditions, defined as $W = \frac{1}{2}(\sigma_{11}\epsilon_{11} + \sigma_{22}\epsilon_{22} + 2\sigma_{12}\epsilon_{12})$;
    \item The fifth channel represents the load constraints in the x-direction. Every node is given the value of the force applied in the x-direction on this load ($0$ if no force is applied on the load);
    \item The sixth channel similarly represents the load constraints in the y-direction;
\end{enumerate}
where $(\sigma_{11}, \sigma_{22}, \sigma_{12})$ and $(\epsilon_{11}, \epsilon_{22}, \epsilon_{12})$ are respectively the components of the stress and strain fields.

We randomly selected a combination of conditions (volume fraction, boundary conditions, loads) to generate every structure and then computed the optimal topology using the SIMP-based TO library ToPy~\cite{Hunter2007william}.
We defined the possible conditions in a similar way to what was done in previous works, namely:
\begin{enumerate*}
    \item The volume fraction is chosen in the interval $[0.3, 0.5]$, with a step of $0.02$;
    \item The displacement boundary conditions are chosen among 42 scenarios for training and five additional scenarios only used for testing;
    \item The loads are applied on unconstrained nodes randomly selected on the domain's boundary. The direction is selected in the interval $[0, \pi]$, with a step of $\frac{\pi}{6}$.
\end{enumerate*}

The main dataset is divided into training, validation, and testing as follows: 
\begin{enumerate}
    \item The \textbf{training data} consist of 30,000 combinations of constraints containing 42 of the 47 boundary conditions;
    \item The \textbf{validation data} consist of 200 new combinations of constraints containing the same 42 boundary conditions;
    \item The \textbf{\emph{level 1} test data} consist of 1800 new combinations of constraints containing the same 42 boundary conditions;
    \item The \textbf{\emph{level 2} test data} consist of 1000 new combinations of constraints containing five out-of-distribution boundary conditions.
\end{enumerate}

In all test data, the combination of constraints is unseen. While \textit{level 1} dataset contains boundary conditions that are also in the training data, we introduce more difficult conditions in \textit{level 2} to rigorously compare the TopoDiff model's generalization ability with existing methods.
In addition to the main dataset, two other datasets consisting of 12,000  and 30,000 non-optimal structures are used to train
\textit{regressor} and \textit{classifier guidance} models.

\subsection{Evaluation Metrics}\label{metrics}
Using relevant evaluation metrics is vital for every scientific study. It is particularly critical for mechanical design generation because most metrics used in DGMs do not correspond to the physical objective one wants a design to achieve.
In this work, contrary to most generative models applied to TO in previous works, we do not use pixel-wise error as a final evaluation metric because it does not guarantee low compliance, which is the objective we are trying to achieve.

Hence, we define and use four evaluation metrics that reflect the compliance minimization objective, as well as the constraints that the generated structures have to respect:
\begin{enumerate}
    \item Compliance error (CE) relative to the ground truth, defined as:
    $CE = {(C(\hat{y}) - C(y))}/{C(y)}$
    where $C(y)$ and $C(\hat{y})$ are, respectively, the compliance of the SIMP-generated topology and the topology generated by our diffusion model.
    It should be noted that a negative compliance error means that our model returns a topology with lower compliance than the ground truth;
    \item Volume fraction error (VFE) relative to the input volume fraction, defined as:
    $VFE = |VF(\hat{y}) - VF(y)|/VF(y)$
    where $VF(y)$ and $VF(\hat{y})$ are, respectively, the prescribed volume fraction and the volume fraction of the topology generated by our diffusion model;
    \item Load violation (LV), defined as a boolean that is 1 if there is no material at a place where a load is applied and 0 if there is always material where loads are applied;
    \item Presence of floating material (FM), defined as a boolean that is 1 if the topology contains floating material and 0 otherwise.
\end{enumerate}

A model which generates samples with high scores on these metrics is expected to yield high-performance manufacturable designs.

\subsection{Choice of hyperparameters}\label{choice_scale}
One of the most crucial hyperparameters is the gradient scales in our guidance strategies.
These parameters quantify the relative importance of compliance minimization and floating material avoidance.
As explained in Sec. \ref{dataset}, a validation dataset of 200 structures was used to perform hyperparameter tuning. We used a grid search method to decide the hyperparameters using compliance error and floating material presence as evaluation metrics. Topology generation and FEA were used to evaluate the results.

\section{Results and Discussions}\label{results}

\begin{table*}[h!]
\centering
\resizebox{\textwidth}{!}{%
\begin{tabular}{@{}c|ccc|ccc@{}}
                  \multicolumn{1}{c}{} & \multicolumn{3}{c}{\textbf{\textit{Level 1} test data}}            & \multicolumn{3}{c}{\textbf{\textit{Level 2} test data}}             \\ \midrule
\textit{Model}                              & TopologyGAN          & Unguided TopoDiff & Guided TopoDiff      & TopologyGAN          & Unguided TopoDiff & Guided TopoDiff           \\ \midrule
\textbf{Average Compliance Error (\%)}                    & 48.51 $\pm$ 16.38         & \textbf{4.10} $\pm$ 0.88    & 4.39 $\pm$ 0.94         & 143.08 $\pm$ 38.50       & 22.13 $\pm$ 8.52           & \textbf{18.40} $\pm$ 5.88      \\
\textbf{Median Compliance Error (\%)}                     & 2.06          & \textbf{0.80}    & 0.83          & 6.82          & 1.88             & \textbf{1.82}     \\
\textbf{Proportion of Compliance Error \textgreater 30\% (\%)}                     & 10.11          & \textbf{2.33}    & 2.56          & 24.10          & 8.20             & \textbf{8.10}     \\
\textbf{Average Volume Fraction Error (\%)}                   & 11.87 $\pm$ 0.52        & 1.86 $\pm$ 0.03            & \textbf{1.85} $\pm$ 0.03         & 14.31 $\pm$ 0.75        & 1.81 $\pm$ 0.04            & \textbf{1.80} $\pm$ 0.04        \\
\textbf{Proportion of Load Violation (\%)}                   & \textbf{0.00} & \textbf{0.00}    & \textbf{0.00} & \textbf{0.00} & \textbf{0.00}    & \textbf{0.00} \\
\textbf{Proportion of Floating Material (\%)}                   & 46.78         & 6.64             & \textbf{5.54} & 67.90         & 7.53             & \textbf{6.21}     \\ \bottomrule
\end{tabular}%
}
\caption{Comparison of performance between TopologyGAN and TopoDiff (guided and not guided) on the two level test sets. Values after $\pm$ indicate the 95 \% confidence interval around averages. The values in bold are the best ones for each level.}
\label{results_vs_cGAN}
\vspace{-2mm}
\end{table*}

\subsection{Evaluation of the full diffusion model}\label{eval_full_model}
To evaluate the performance, we use the two test sets described in Sec. \ref{dataset}, corresponding to two difficulty levels.
We run every test nine times and then compute the results' average.
We compare the performance of our model on all evaluation metrics (Sec. \ref{metrics}) to a state-of-art cGAN model, named TopologyGAN~\cite{Nie2021}, which performs the same task as our model.

\begin{figure}[h]
  \centering
  \includegraphics[scale = 0.92]{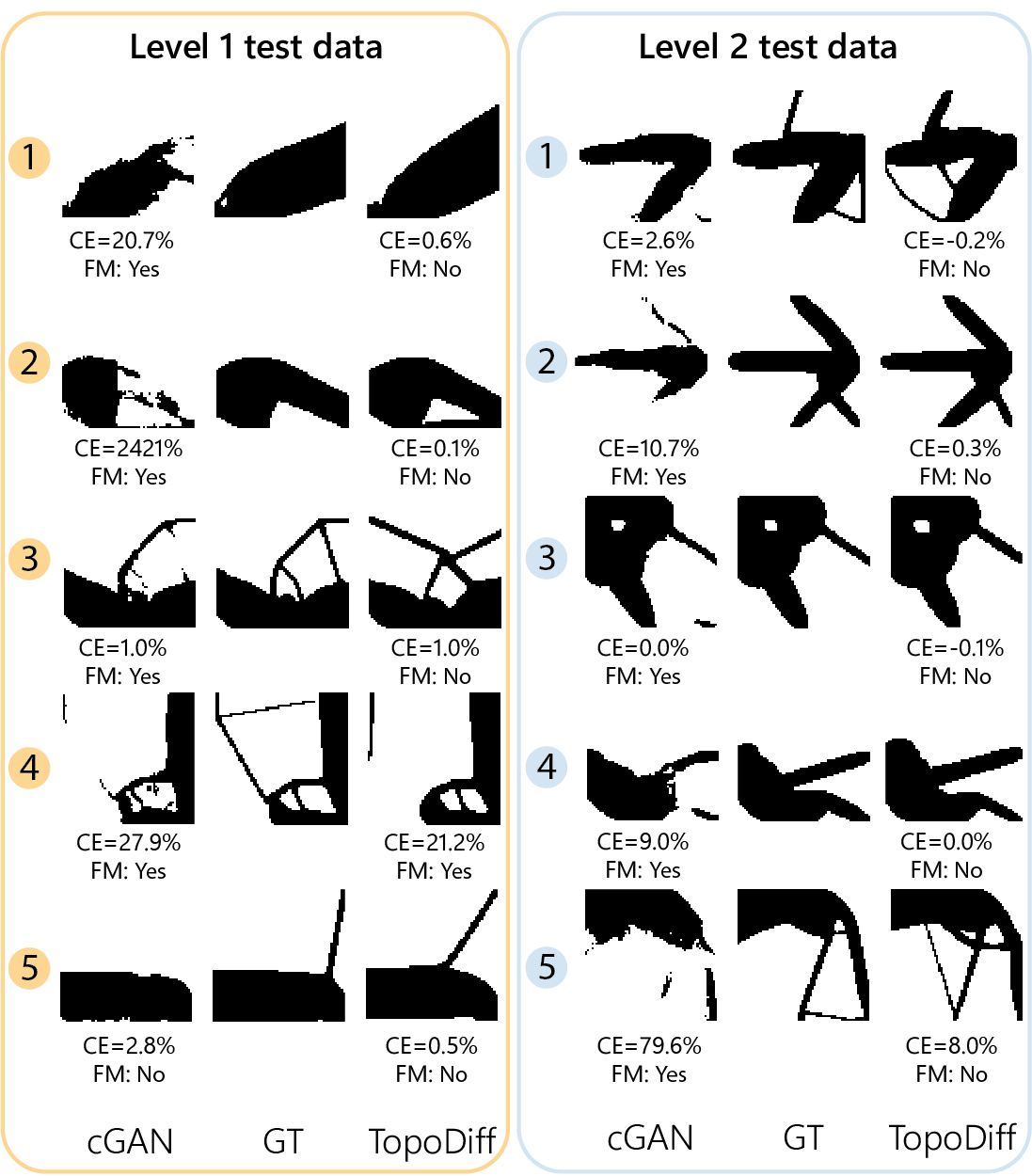}
  \caption{Comparison of generated structures on randomly selected samples from both test datasets. GT stands for ground truth, CE is the compliance error relative to the GT, and FM indicates the presence or not of floating material.}
  \label{fig:generated_examples}
\end{figure}

Fig. \ref{fig:generated_examples} shows examples of a few structures obtained with the SIMP method (ground truth), with TopologyGAN, and with TopoDiff for randomly selected constraints from \textit{level 1} and \textit{level 2} test sets.
Qualitatively, we notice that TopologyGAN tries to mimic pixel-wise the topology obtained from SIMP but neglects both the compliance and the manufacturability of the generated structures, which almost all have some floating material and high compliance error.
The ten structures generated by TopoDiff, on the other hand, may visually differ more from the SIMP results but have better physical properties than TopologyGAN. Only one of the TopoDiff-generated structures have no floating material, and all ten outperform the TopologyGAN structures in terms of compliance error. To confirm these qualitative observations, Table \ref{results_vs_cGAN} summarizes the performance of the structures obtained with all test sets. TopoDiff outperforms TopologyGAN on all the metrics.

On the \textit{level 1} test set, TopoDiff notably reduces the average CE by a factor of eleven and the proportion of FM by more than a factor of eight.
The proportion of non-manufacturable designs thus drops from 46.8\% with cGAN to 5.5\% with TopoDiff.
It also significantly reduces the average VFE from 11.9\% to 1.9\%.

On the \textit{level 2} test set, TopoDiff achieves demonstrates strong generalizability performance.
It performs an eight-times reduction in the average CE, from 143.1\% to 18.4\%, and a four-times reduction in the median CE.
Non-manufacturability drops from 67.9\% to 6.2\%, while the VFE is reduced by a factor of eight, from 14\% to less than 2\%.
A paired one-tailed t-test confirms a reduction of the average CE and of the average VFE with a p-value of $9 \cdot 10^{-12}$ and $5 \cdot 10^{-160}$ respectively. These results show the efficacy of diffusion models in learning to generate high-performing and manufacturable structures for a wide set of testing conditions.

\subsection{Efficiency of guidance strategy}\label{efficiency_guidance}
\subsubsection{Surrogate models}\label{surrogate}
Guidance can only work if the regressors and classifiers can perform well on the challenging task of predicting compliance and floating material for noisy images.
Table \ref{perf_reg_class} shows the compliance regressor and floating material classifier performance according to the noise level.
These results show that both surrogate models are very reliable on low-noise structures, and as expected, their performance decreases with an increase in noise.

\begin{table}[h!]
\resizebox{\columnwidth}{!}{%
\begin{tabular}{@{}c|cccc@{}}
\toprule
                          & 0-25\% noise & 25-75\% noise & 75-100\% noise & Global \\ \midrule
\textbf{Regressor R2 (\%)}     & 82.4      & 82.4       & 61.8       & 77.3    \\
\textbf{Classifier accuracy (\%)} & 98.8         & 76.8          & 54.6          & 76.8       \\ \bottomrule
\end{tabular}
}
\caption{Performance (R2-score and accuracy) of both surrogate models on validation data with respect to noise level.}
\label{perf_reg_class}
\end{table}
\vspace{-4mm}

\subsubsection{Ablation study}\label{ablation}
To evaluate the impact of our guidance strategy on the performance of TopoDiff, we also tested it \textit{without} guidance on the two test sets, as shown in Table \ref{results_vs_cGAN}.

With \emph{in-distribution} boundary conditions (\textit{level 1}), our guidance strategy has no significant impact on compliance error (both on average and median).
A two-tailed paired t-test does not reject the null hypothesis ($p = 0.1$).
We believe that this happens because the diffusion model has implicitly learned to perform well with these boundary conditions and does not need explicit compliance guidance.
In contrast, our guidance strategy significantly impacts the proportion of floating material, with decreases from $6.6\%$ to $5.5\%$.

With \emph{out-of-distribution} boundary conditions (\textit{level 2}), the positive impact of our guidance strategy is evident both on the average compliance error and on the proportion of floating material.
A paired one-tailed t-test confirms a reduction of the average CE with a p-value of $0.05$.
The average compliance error is reduced by $17\%$ and the average proportion of floating material by $18\%$.
As expected, guidance seems to have no effect on load respect and on volume fraction error. 
More interestingly, guidance seems to have no significant effect on the median of the compliance error, which suggests that compliance \textit{regressor guidance} primarily reduces the number of structures with very high compliance errors.

\subsection{Limitations and future work}\label{limitations}
TopoDiff shows good performance and good generalization to out-of-distribution boundary conditions.
The proposed guidance strategy is beneficial to minimizing compliance and ensuring constraints, such as manufacturability.
However, several challenges still need to be addressed.
The most significant limitation is the computation time, diffusion models being slower than GANs.
It takes 0.06 seconds for TopologyGAN to generate one topology, while TopoDiff needs 21.59 seconds.
Reducing the computation time of diffusion models has recently seen significant successes~\cite{speed_DM1}, which will directly improve TO-based diffusion models.
Other potential directions for future research include applying TopoDiff to more complex TO problems, notably 3D problems, and scaling it to
higher resolutions and more boundary conditions. Reducing dependency on mesh size and large training datasets is also critical. We provide a framework for conditioning a diffusion model with constraints, training it on optimal data, and guiding it with a regressor predicting physical performance and some classifiers predicting the respect of constraints. This
is a general method that should allow for solving similar design generation problems involving performance objectives and constraints, such as aerodynamic design~\cite{pcdgan} and bicycle synthesis~\cite{regenwetter2022biked}.
Future work also includes expanding the TopoDiff framework to solve many inverse problems in engineering domains with multi-modal inputs.

\section{Conclusion}\label{conclusion}
Diffusion Models have been extremely successful in modeling high-dimensional multi-modal distributions with astonishing results in high-fidelity image generation.
We propose TopoDiff --- a conditional diffusion model to perform end-to-end topology optimization. By introducing conditional diffusion models for topology optimization, we show that diffusion models can outperform GANs in engineering design applications too.
Our model is augmented with an explicit guidance strategy to ensure performance maximization and avoidance of non-manufacturable designs. TopoDiff achieves an eight-times reduction in the average compliance error and produces 11-times fewer non-manufacturable designs compared with a state-of-art conditional GAN. It also achieves an 
eight-times reduction in volume fraction error and generalizes well to out-of-distribution boundary conditions. 
More generally, we provide a diffusion model-based framework to solve many physical optimization problems in engineering with performance objectives and constraints.

\section*{Acknowledgments}
The authors gratefully acknowledge the funding received from Mines Paris Foundation. They would also like to thank MISTI France for supporting this research. 

\bibliography{bibliography}

\clearpage

\renewcommand{\thesection}{A\arabic{section}}
\renewcommand{\thetable}{A\arabic{table}}
\renewcommand{\thefigure}{A\arabic{figure}}
\setcounter{section}{0}
\section{Additional datasets used for surrogate models}\label{add_datasets}
The main dataset used to train, validate and test the diffusion model was presented in the paper.
This section provides more information about the two other datasets that were used to train and validate the two surrogate models, namely the regressor predicting compliance and the classifier predicting the presence of floating material.

For the regressor predicting compliance, we used a dataset of 72000 labeled samples, with the label indicating the compliance of the topology under the given constraints.
Every sample is an image containing eight channels (black and white topology, volume fraction, von Mises stress, strain energy density, load in x-direction, load in y-direction, x-displacement boundary condition and y-displacement boundary condition).
In order to give our regressor good generalization abilities, we not only used optimal topologies (generated by SIMP), but also negative samples, corresponding to non optimal topologies.
This dataset includes:
\begin{itemize}
    \item 30000 images from the diffusion model main dataset, corresponding to \textit{optimal topologies} (25000 for training, 5000 for validation);
    \item 12000 images corresponding to \textit{non-optimal topologies}, generated using the \textit{fake-load method}, explained below (10000 for training, 2000 for validation);
    \item 30000 images corresponding to \textit{non-optimal topologies}, generated by the unguided conditional diffusion model by using the same constraints as in the main dataset (25000 for training, 5000 for validation).
\end{itemize}
It should be noted that the final dataset contains less than 72000 samples because it was filtered to remove outlier structures (with compliance higher than 50 for the first 42000 and with compliance higher than 25 for the diffusion-generated structures).
To generate non-optimal structures, we notably used the \textit{fake-load method}, which consists in adding an extra load to the input constraints given to SIMP for generating the structure. SIMP thus generates a topology that is optimal for the extra-loaded constraints but not for the real ones. This method allows to generate non-optimal data while still respecting basic requirements like the presence of material where loads are applied.

For the classifier predicting the presence of floating material, we used a dataset of 70000 labeled samples (58000 for training and 12000 for validation), with the label being a boolean indicating if floating material is present on the topology or not.
Every sample is an image containing only one channel: the black and white topology.
This dataset includes:
\begin{itemize}
    \item the first 15000 images from the diffusion model main dataset, on which floating material was added;
    \item the first 14000 images from a dataset containing topologies at diverse volume fractions, on which floating material was added;
    \item the first 6000 images from the non-optimal topologies dataset generated with the \textit{fake-load method}, on which floating material was added;
    \item the next 15000 images from the diffusion model main dataset, on which floating material was \emph{not} added;
    \item the next 14000 images from a dataset containing topologies at diverse volume fractions, on which floating material was \emph{not} added;
    \item the next 6000 images from the non-optimal topologies dataset generated with the \textit{fake-load method}, on which floating material was \emph{not} added.
\end{itemize}

It should be noted that while some datasets were used to train several different models, we have paid careful attention to avoiding leakage of data from training datasets to validation datasets.
All models were validated on data that they had not been trained on.
The final TopoDiff model (diffusion model + surrogate models) was tested on data that was never used for training or validating any of its component models.

\end{document}